\documentclass{article}
\usepackage{microtype}
\usepackage{graphicx}
\usepackage{subfigure}
\usepackage{booktabs}
\usepackage{placeins}
\usepackage{arydshln}
\usepackage[export]{adjustbox}
\usepackage{hyperref}

\usepackage[accepted]{synsml2023}

\usepackage{amsmath}
\usepackage{amssymb}
\usepackage{mathtools}
\usepackage{amsthm}

\usepackage[capitalize,noabbrev]{cleveref}

\usepackage{balance}

\usepackage[textsize=tiny]{todonotes}

\synsmltitlerunning{Integrating processed-based models and machine learning for crop yield prediction}

\begin{document}

\twocolumn[
\synsmltitle{Integrating processed-based models and machine learning \\ for crop yield prediction}
\synsmlsetsymbol{equal}{*}

\begin{synsmlauthorlist}
\synsmlauthor{Michiel G.J. Kallenberg}{equal,WUR}
\synsmlauthor{Bernardo Maestrini}{equal,WUR}
\synsmlauthor{Ron van Bree}{WUR}
\synsmlauthor{Paul Ravensbergen}{WUR}
\synsmlauthor{Christos Pylianidis}{WUR}
\synsmlauthor{Frits van Evert}{WUR}
\synsmlauthor{Ioannis N. Athanasiadis}{WUR}
\end{synsmlauthorlist}

\synsmlaffiliation{WUR}{Wageningen University \& Research, the Netherlands}

\synsmlcorrespondingauthor{Michiel Kallenberg}{michiel.kallenberg@wur.nl}

\synsmlkeywords{crop growth modeling, hybrid modeling, metamodel, machine learning, transfer learning, potato, agriculture}

\vskip 0.3in
]

\printAffiliationsAndNotice{\synsmlEqualContribution}

\begin{abstract}
Crop yield prediction typically involves the utilization of either \textit{theory-driven} process-based crop growth models, which have proven to be difficult to calibrate for local conditions, or \textit{data-driven} machine learning methods, which are known to require large datasets. In this work we investigate potato yield prediction using a hybrid meta-modeling approach. A crop growth model is employed to generate synthetic data for (pre)training a convolutional neural net, which is then fine-tuned with observational data. When applied in silico, our meta-modeling approach yields better predictions than a baseline comprising a purely data-driven approach. When tested on real-world data from field trials (n=303) and commercial fields (n=77), the meta-modeling approach yields competitive results with respect to the crop growth model. In the latter set, however, both models perform worse than a simple linear regression with a hand-picked feature set and dedicated preprocessing designed by domain experts. Our findings indicate the potential of meta-modeling for accurate crop yield prediction; however, further advancements and validation using extensive real-world datasets is recommended to solidify its practical effectiveness.
\end{abstract}

\section{Introduction}
Crop yield prediction plays an important role in ensuring food security, optimizing agricultural practices, and managing risks in the farming industry. By accurately estimating the expected crop yields, stakeholders ranging from farmers and policymakers to traders and researchers can make informed decisions and take proactive measures. In a world facing increasing population growth and climate uncertainties, the ability to predict crop yields has become indispensable for addressing global food challenges and fostering sustainable agricultural practices.

In crop yield prediction, \textit{theory-driven} process-based crop models and \textit{data-driven} machine learning models are usually seen as opposite approaches \cite{Leng20,Paudel2022,Maestrini2022}. In this work, we investigate the use of a meta-modeling approach to combine the best of these two worlds. The integration of the two approaches may help overcome their individual approach limitations, for process-based models the limited processes included in the model \cite{Itte13}, and their difficulty in adapting to local conditions, and for data-driven models the need for large, variable and orthogonal datasets, that are often unavailable in the agricultural context. 

One common method of integrating data-driven and process-based approaches is through the development of a metamodel \cite{Wall19, Pylianidis2022}. The workflow to create a metamodel is typically the following: a synthetic dataset is created using a process-based model fed with different inputs representative for the conditions where the model will be used, and then a data-driven model is fit on the synthetic dataset \cite{Raza12, Pylianidis2022}. Usually only a subset of the input (e.g. weather, management, soil texture) and a subset of the output (e.g. yield, or yield and leached N) is retained in the metamodel.

To align model predictions with local conditions, crop yield prediction using process-based crop growth models approaches typically involves a tedious calibration process \cite{Seid18}. Calibration of these models can be challenging due to the complexity of cropping systems, spatial and temporal variability, limited data availability, and parameter sensitivity. Calibrating a data-driven method is often considered easier compared to calibrating a process-based model. For data-driven models transfer learning techniques, such as fine-tuning, can be leveraged to adapt a generic model to local conditions. 

\begin{figure*}[t]
\centering\includegraphics[width=1.5\columnwidth]{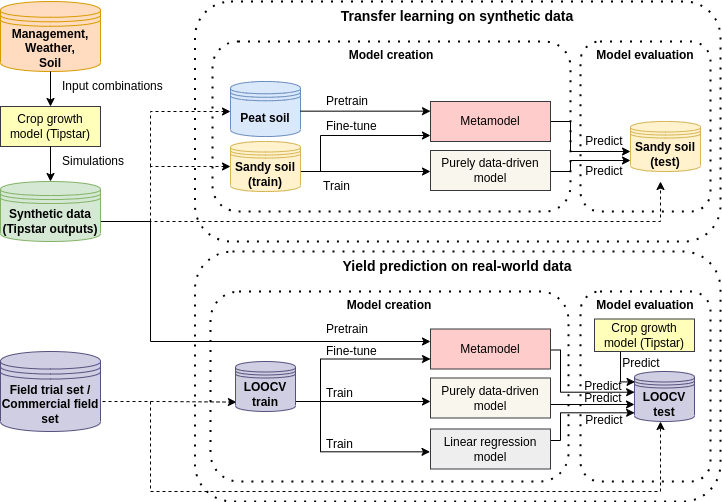}
\caption{Schematic overview of experimental setup, data flow and model development}
\label{fig:workflow}
\end{figure*}

In this work, we compare three methods for predicting potato yield, quantified as fresh tuber weight at harvest time: (1) a process based crop growth model, in our case Tipstar \cite{Jans08} (2) a (fine-tuned) metamodel, pretrained with synthetic data generated by Tipstar, and (3) a purely data-driven model, not (pre)trained with Tipstar. 

\section{Materials and methods}
A schematic overview of the experimental setup, data flow and model development is shown in Figure \ref{fig:workflow}.
\subsection{Models}
\subsubsection{Process-based crop growth model}
Tipstar \cite{Jans08} is a crop growth model that simulates potato growth under water and nitrogen limited conditions with daily time steps. The input to the model are crop management (planting, fertilization, and irrigation), weather (solar radiation, rainfall, maximum and minimum air temperature, wind), and soil characteristics (texture, organic matter percentage, van Genuchten parameters \cite{Genu80}). The model was not calibrated for this specific setup (i.e., no model parameters were adjusted to fit the data of this study), but we re-used default parameters that were derived from historical data from the same country (i.e., the Netherlands).

\subsubsection{Metamodel}
We developed a metamodel of Tipstar by creating a synthetic dataset of 86,000~entries, which was then used to pretrain a convolutional neural network. The synthetic dataset was created by taking the full factorial of the model inputs weather (7 locations, 32 years) and soil (32 soil types \cite{Wost13}). For each of the resulting 7,168 combinations we generated 12 simulations using random sampling while varying the following model inputs: nitrogen fertilization, sowing date, irrigation, maximum rooting depth, and cultivar earliness.

The metamodel was constructed as a multi-stream (n=3) convolutional neural network. The first stream processed the temporal data (i.e. solar radiation, rainfall, maximum and minimum air temperature, cumulative irrigation and cumulative nitrogen input) with twice a 1D convolution (number of filters 20 and 7, kernel sizes 3 and 2) followed by an average pooling (5 and 5). The second stream processes the scalars (i.e. maximum rooting depth, sowing day of year and cultivar earliness) with two dense layers (20, and 20). The (optional) third stream processes the soil characteristics (i.e., clay, loam, organic matter percentage, soil moisture at saturation, and the Van Genuchten parameters: $\alpha$, $\lambda$ and $n$ \cite{Genu80}) with a 1D convolution (number of filers 5, kernel size 5) followed by an average pooling (24). The three streams are all flattened and concatenated, and subsequently fed to three dense layers (25, 5 and 1).

\subsubsection{Data-driven model}
As a baseline for our model comparisons we trained a purely data-driven model. This model was not pretrained with synthetic data, but was trained with observational data. The model had the same architecture as the metamodel.

(Pre)training of the metamodel and the data-driven model was done with mean squared error (MSE) as a loss function. We used the ADAM optimizer with an initial learning rate of 0.001. Training was monitored on an hold-out validation set. To prevent overfitting, we used early stopping (min\_delta=0.001, patience=20), and learning rate reduction (factor=0.5, min\_delta=0.001, patience=10).

\subsection{Experiments}
We conducted two experiments to investigate the merits of our meta-modeling approach. 

\subsubsection{Transfer learning on synthetic data}
The first experiment served as a proof-of-concept of our transfer learning approach, and was done with synthetic data only. We selected "soil" as our domain of interest, since, in general, potato yield depends substantially on the characteristics of the soil \cite{Port99}. We divided the synthetic dataset in two subsets; the first set (i.e., source domain) contained the simulations on peat soils (category 2xx in BOFEK 2012 \cite{Wost13}) and the second (i.e., target domain) the sandy soils (category 3xx in BOFEK 2012 \cite{Wost13}). We confirmed that yields differed between the two soil types, which was found to be partially caused by a different yield response to water availability.

A metamodel (excluding the soil input stream) was pretrained on the peat soil dataset (n=10k), and then fine-tuned with the sandy soil dataset. For the fine-tuning we chose to freeze all layers of the network, except for the last two layers, as this approach yielded slightly superior results compared to alternatives such as training all layers, and training the first layers only. For comparison, we trained a data-driven model with the same architecture of the metamodel. The data-driven model was exclusively trained with the sandy soil dataset. Both models were evaluated on a hold-out set (n=17k) with sandy soils.

Transfer learning is especially instrumental in cases where there is a limited availability of data from the target domain. We simulated such a scenario by limiting the size of the dataset available for fine-tuning; we evaluated three different levels of size n=50, 200, and 1000 respectively.

\subsubsection{Yield prediction on real-world data}
The second experiment was done with two real-world datasets; one from field trials and one from commercial fields. The field trial dataset (n=303, 1994-2003) came from a collection of 36 potato experiments carried out by Wageningen University and Research. The factors that were varied in the experiments were cultivar, nitrogen fertilization rates and timings, irrigation rates and timings, sowing date, and planting density. The field trials were conducted under well-controlled conditions, with e.g. effective pest management. The commercial field dataset (n=77, 2015-2020) was collected from an arable farm located near Wageningen University and Research. As the primary focus of the commercial cultivation was profit rather than scientific research, crop management was carried out based on standard practices aiming to achieve an optimal balance between yield and costs.

A metamodel was pretrained on the synthetic dataset, and subsequently fine-tuned with the observational dataset. To prevent data leakage, we excluded the years in which the experiments of the potato trials dataset were performed from the synthetic dataset. For comparison, we trained a data-driven model with the same architecture of the metamodel. Because the observational dataset was relatively modest in size, we used leave-one-out cross-validation for model development and testing. As weather is a major determinant of final yield, we used weather years as criteria for the splits. To assess the stability of the training, we replicated the trainings three times using different random seeds.

As an additional baseline, representing one of the most commonly used data-driven approaches in yield prediction, we trained a simple linear regression model with a hand-picked feature set (i.e. earliness, sowing date, precipitation, and average daily temperature) selected by domain experts. The time series variables (i.e., precipitation, and average daily temperature) were preprocessed by averaging over the period from May-August.

\section{Results}
\begin{figure}[h]
\begin{center}
\includegraphics[width=0.85\columnwidth]{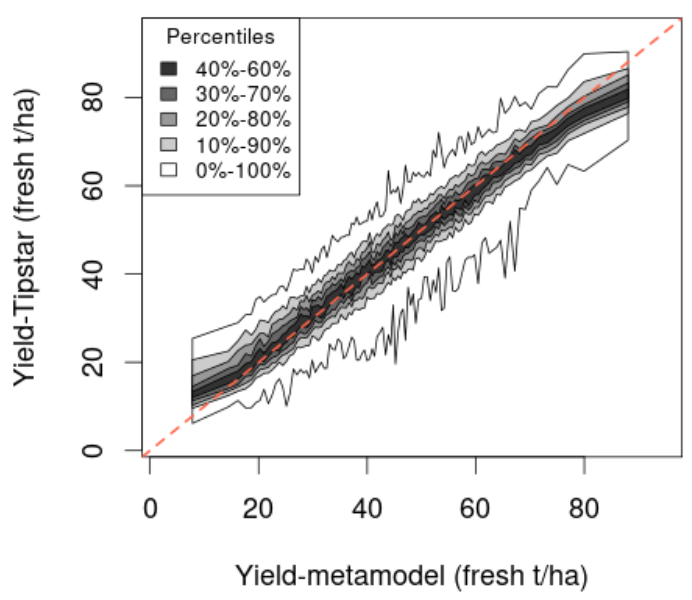}
\caption{Fitness of the metamodel on the (hold-out) synthetic dataset (n=13,000; RMSE=5.1, black}{$r=0.95$})
\label{fig:metamodel-synthetic}
\end{center}
\end{figure}

We (pre)trained a metamodel for potato yield prediction, using synthetic data obtained from the process-based crop growth model Tipstar. The obtained metamodel was able to reproduce Tipstar with an RMSE of 5.1 fresh tonne/ha and $r=0.95$ on a hold-out set (n=13,000) (see Fig.\ref{fig:metamodel-synthetic}).

\subsection{Transfer learning on synthetic data}
In our first experiment we investigated the merits of incorporating fine-tuning in the training process of our metamodel, to cover for domain shifts. As an alternative approach we included a purely data-driven baseline that was not pretrained with the crop growth model.

\begin{table}[b]
\caption{Model performance as function of fine-tuning set size}
\label{fine-tuning}
\begin{center}
\begin{footnotesize}
\begin{sc}
\begin{tabular}{llccccc}
\toprule
metric & model & \multicolumn{4}{l}{fine-tuning set size}\\
& &  \textit{0$^a$} & 50 & 200 & 1,000\\
\midrule
$r$ & Metamodel & \textit{0.884} & 0.899 & 0.905 & 0.908 \\
& Data-driven$^b$ & N/A & 0.767 & 0.790 & 0.902 \\
\hdashline
RMSE & Metamodel & \textit{10.3} & 7.6 & 7.4 & 7.2 \\
& Data-driven$^b$ & N/A & 11.2 & 11.0 & 7.5 \\
\bottomrule
\multicolumn{7}{l}{$^a$\footnotesize{\textit{Exclusively pretrained (on source domain)}}} \\
\multicolumn{7}{l}{$^b$\footnotesize{\textit{Exclusively trained on fine-tuning set}}} \\
\end{tabular}
\end{sc}
\end{footnotesize}
\end{center}
\end{table}

Table \ref{fine-tuning} reports on the performance of the models as function of the fine tuning set size. When target domain data availability is limited, best results are obtained with the fine-tuned metamodel. These results indicate that (1) fine-tuning improves the performance of a pretrained metamodel, and (2) pretraining with synthetic data obtained from a processed based crop model is effective, even when that data is collected from a source domain that differs from the target domain. 

\subsection{Yield prediction on real-world data}
In our second experiment we evaluated our metamodel approach on real-world data.

\begin{figure}[h]
\begin{center}
\centerline{\includegraphics[width=\columnwidth]{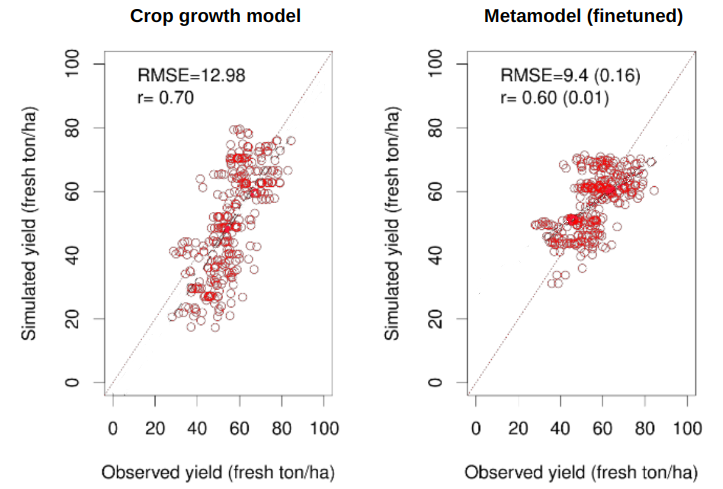}}
\caption{Fitness of the crop growth model and the metamodel on the field trial set (n=303). The number in parentheses represents the standard deviation of the three replicates of the metamodel trainings.}
\label{trials}
\end{center}
\end{figure}

\begin{figure}[t]
\begin{center}
\centerline{\includegraphics[width=\columnwidth]{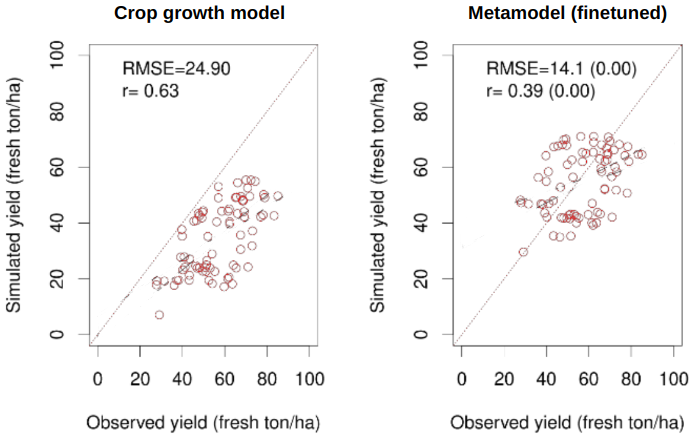}}
\caption{Fitness of the crop growth model and the metamodel on the commercial field set (n=77). The number in parentheses represents the standard deviation of the three replicates of the metamodel trainings.}
\label{commercial}
\end{center}
\end{figure}

Figures \ref{trials} and \ref{commercial} show scatter plots of the predicted and the observed yields for both the crop growth model and the metamodel on the field trial set (n=303), and the commercial field set (n=77), respectively. In general, predictions of both models are better in the field trial set than in the commercial field set. While for the field trial set the predictions of the crop growth model are, on average, in line with the observed yields, for the commercial fields the crop growth model systematically underestimates the yield. This emphasizes the necessity of calibrating a crop growth model prior to its application to real-world data. The (fine-tuned) metamodel, on the contrary, exhibits no systematic bias, in either sets. Yet, the metamodel, has a relatively high scatter.

\begin{table}[b]
\caption{Model performance on real-world datasets}
\begin{center}
\begin{footnotesize}
\begin{sc}
\begin{tabular}{llll}
\toprule
metric & model & Trial & Commercial\\
\midrule
$r$ & Crop growth model & 0.70 & 0.63 \\
& Metamodel & 0.60 & 0.39\\
& Data-driven & 0.02 & 0.35\\
& Linear regression & 0.64 & 0.72\\
\hdashline
RMSE & Crop growth model & 12.98 & 24.90\\
& Metamodel & 9.40 & 14.10\\
& Data-driven & 15.80 & 14.10\\
& Linear regression & 8.93 & 9.75\\
\end{tabular}
\end{sc}
\end{footnotesize}
\end{center}
\end{table}

As a reference, we trained a purely data-driven model (i.e., we did not use the crop growth model for pretraining), with the same architecture of the metamodel. With an RMSE=15.80, and r=0.02 on the field trial set, and RMSE=14.10, and r=0.35 on the commercial set, the purely data-driven model showed poor performance. Arguably the complexity of the model was excessive considering the limited number of available training samples. We also trained a significantly less complex model, specifically a linear regressor, for which the feature set was hand picked by domain experts, with dedicated preprocessing. This model yielded an RMSE=8.93, and r=0.64 on the field trial set, and RMSE=9.75, and r=0.72 on the commercial field set, outperforming both the crop growth model and the metamodel on the latter dataset.

\section{Discussion and conclusion}
In this work, we investigated potato yield prediction using a meta-modeling approach that integrates a \textit{theory-driven} process-based crop growth model into a \textit{data-driven} machine learning method. We found that, in silico, our metamodel yields better predictions than a purely data-driven approach. When tested on real-world data, comprising a field trial set and a commercial field set, the metamodel yields competitive results with respect to the crop growth model. In the latter set, which was characterized by a small sample size, however, a simple linear regression model with a hand-picked feature set, representing one of the most commonly used data-driven approaches in yield prediction, outperformed both the crop growth model and the metamodeling approach.

Our results indicate a benefit of pretraining a machine learning model with synthetic data obtained from a crop growth model, rather than using real-world data only. In our in silico experiments, pretrained models fine-tuned with only 50 data points had similar performance to data-driven models using 1,000 data points. This is important for data poor domains as agriculture. Also, training data exclusively obtained from a real-world setting may lack important contrasts in the input space, as, for example, especially in commercial settings, management practices are typically standardised. In this context, employing synthetic data may be seen as a data augmentation strategy that prevents the model from exploiting spurious features, which is especially relevant in models with high complexity. Model over-complexity combined with a small, low-contrast training set, may be one of the reasons why our purely data driven model did not perform well in the real-world datasets. Pretraining our model with synthetic data turned out to be a successfully strategy. It should be noted though that a simple approach, using an expert designed, regression model may still be preferred in a real-world setting.

We conclude that meta-modeling has potential for accurate crop yield prediction, yet further development and validation on large real-world datasets is recommended.

\section{Impact}
Predicting crop yields is crucial for ensuring food security, optimizing agricultural practices, and managing risks in the farming industry. In crop yield prediction, theory-driven process-based crop models and data-driven machine learning models are usually seen as opposite approaches. In this work, we explore the utilization of a meta-modeling approach that combines the strengths of both these approaches. By integrating these two methods, we aim to overcome their respective limitations. Process-based models often suffer from a restricted set of included processes, while data-driven models rely on large, diverse, and independent datasets that are often scarce in the agricultural domain. Our findings suggest the potential of meta-modeling for accurate crop yield prediction, and may as such facilitate the uptake of machine learning in agriculture.

\section*{Acknowledgements}
We are grateful to three anonymous reviewers for their insightful comments. This work was partially supported by the Dutch Ministry of Agriculture, Nature and Food Quality project Data driven and High Tech (KB-38-001-002), the Wageningen University and Research Investment Theme \textit{Digital Twins}, and the European Union Horizon Research and Innovation program (Grant  \#\href{https://cordis.europa.eu/project/id/101070496}{101070496}, Smart Droplets).

\FloatBarrier
\balance
\bibliography{main}
\bibliographystyle{synsml2023}
\end{document}